\documentclass{bmvc2k}
\usepackage{graphicx}
\usepackage{amsmath}
\usepackage{amssymb}
\usepackage{booktabs}
\usepackage{bbm}
\usepackage{xcolor}
\usepackage{multirow}
\usepackage{arydshln}
\usepackage{wrapfig}
\usepackage{algorithm2e}
\usepackage{dsfont}
\usepackage{color}
\usepackage{comment}
\usepackage{graphics}

\newcommand{\shadow}[1]{}
\newcommand{\blue}[1]{\textcolor{blue}{#1}}
\newcommand{\red}[1]{\textcolor{red}{#1}}

\newcommand{\brown}[1]{\textcolor{brown}{#1}}
\newcommand{\orange}[1]{\textcolor{orange}{#1}}

\def\b{\blue}
\def\s{\shadow}
\def\r{\red}
\def\o{\orange}

\title{C3: Cross-instance guided Contrastive Clustering}

\addauthor{Mohammadreza Sadeghi}{mohammadreza.sadeghi@mcgill.ca}{1,2}
\addauthor{Hadi Hojjati}{hadi.hojjati@mcgill.ca}{1,2}
\addauthor{Narges Armanfard}{narges.armanfard@mcgill.ca}{1,2}

\addinstitution{
Department of Electrical and Computer Engineering, McGill University\\ 
Montreal, QC, Canada\\
}
\addinstitution{
Mila - Quebec AI Institute\\
Montreal, QC, Canada
}

\runninghead{Sadeghi et al.}{C3:Cross-instance guided Contrastive Clustering}

\def\etal{\emph{et al}\bmvaOneDot}

\begin{document}

\maketitle

\begin{abstract}
Clustering is the task of gathering similar data samples into clusters without using any predefined labels. It has been widely studied in machine learning literature, and recent advancements in deep learning have revived interest in this field. Contrastive clustering (CC) models are a staple of deep clustering in which positive and negative pairs of each data instance are generated through data augmentation. CC models aim to learn a feature space where instance-level and cluster-level representations of positive pairs are grouped together. Despite improving the SOTA, these algorithms ignore the cross-instance patterns, which carry essential information for improving clustering performance. This increases the false-negative-pair rate of the model while decreasing its true-positive-pair rate. In this paper, we propose a novel contrastive clustering method, Cross-instance guided Contrastive Clustering (C3), that considers the cross-sample relationships to increase the number of positive pairs and mitigate the impact of false negative, noise, and anomaly sample on the learned representation of data. In particular, we define a new loss function that identifies similar instances using the instance-level representation and encourages them to aggregate together. Moreover, we propose a novel weighting method to select negative samples in a more efficient way. Extensive experimental evaluations show that our proposed method can outperform state-of-the-art algorithms on benchmark computer vision datasets: we improve the clustering accuracy by $6.6\%$, $3.3\%$, $5.0\%$, $1.3\%$ and $0.3\%$  on CIFAR-10, CIFAR-100, ImageNet-10, ImageNet-Dogs, and Tiny-ImageNet.
\end{abstract}

\vspace{-4mm}
\section{Introduction}
\label{sec:intro}
\begin{wrapfigure}[6]{r}{0.5\textwidth}
\vspace{-26mm}
  \begin{center}
    \includegraphics[width=1\linewidth]{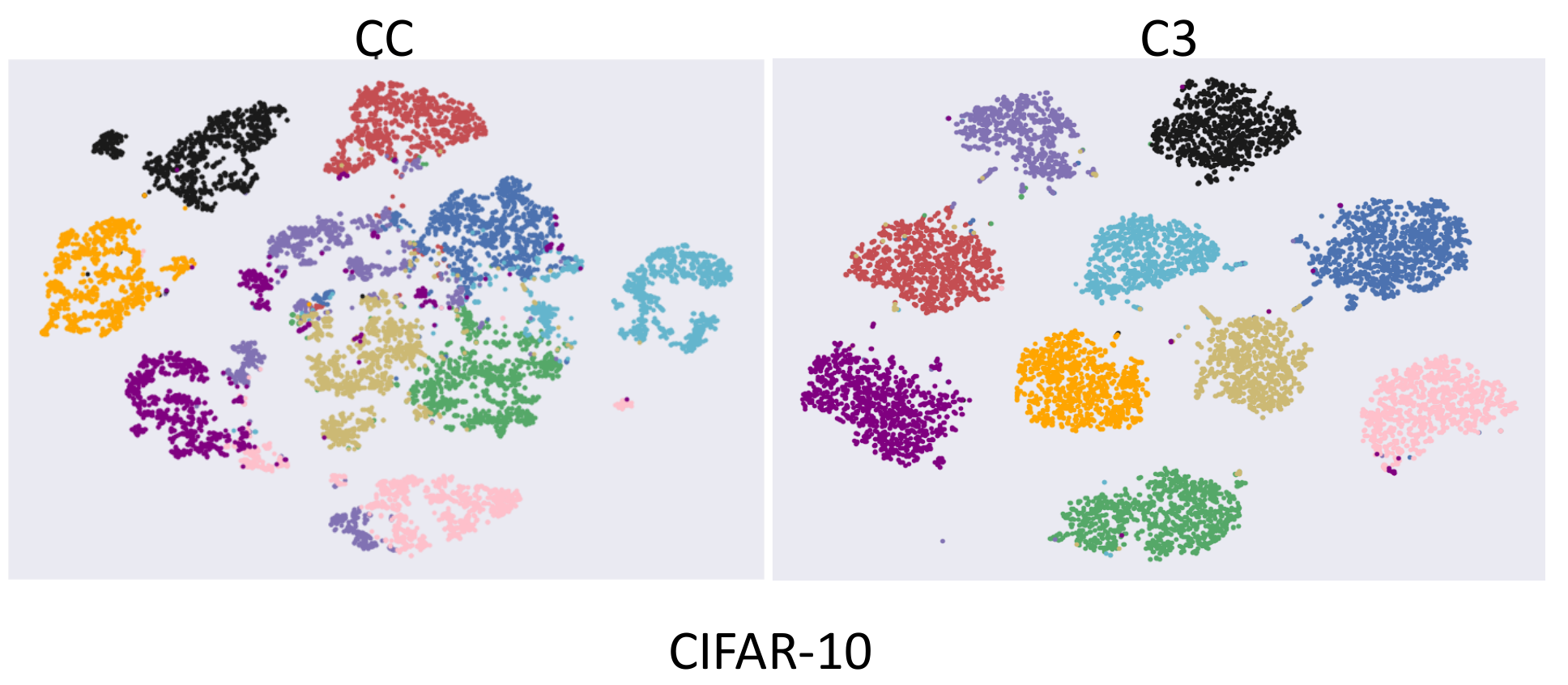}
    \vspace{-10pt}

  \caption{t-SNE visualization of Contrastive Clustering (CC) \cite{CC} and the proposed C3}
    \end{center}
  \label{fig:tsne_cifar}
\end{wrapfigure}
\s{\r{We can Remove the following}
\brown{As a fundamental task in unsupervised machine learning, clustering aims to group similar data points into the same clusters based on a similarity metric in the absence of labels. Clustering algorithms are helpful in many applications, including but not limited to image segmentation \cite{segment}, anomaly detection \cite{anomaly}, medical analysis \cite{medical}, and data retrieval \cite{retriv1,retriv2}, and are often considered an essential pre-processing step in data mining tasks \cite{mining}. It has been an active field of research in machine learning, and a flurry of clustering frameworks has been proposed throughout time. Traditional methods, such as k-means \cite{kmeans} and fuzzy c-means \cite{FCM}, could achieve a promising performance on lower-dimensional datasets \cite{review_old}. Still, they fail to cluster large-scale or high-dimensional data accurately. They assume that input data are already represented by an efficient feature vector, which is generally not valid in the case of high-dimensional data. }}

\s{Following the recent success of deep models in handling unstructured and high dimensional data such as images, researchers have turned their attention to developing deep learning algorithms for clustering \cite{reviewClustermain}. Most early works relied on the power of deep models in learning a generic representation from the input \cite{DEN}. Since this representation is not particularly suited for clustering, this approach yields sub-optimal results. More recently, \textit{deep clustering} methods have been proposed to tackle this issue \cite{DEC}. They jointly train a representation with the clustering task to learn a feature space in which the data can be more explicitly clustered into different groups. As a result, these models managed to achieve a reliable performance on complex computer vision datasets. }

In the past few years, self-supervised learning \cite{Rotate,lin2021inter,tiwari2021self}, particularly contrastive learning, has established itself as a state-of-the-art representation learning algorithm \cite{SimCLR}. They generate a transformed version of the input samples and attempt to learn a representation in which augmentations of the same data point are closer together and further away from other points. They managed to outperform other baselines, particularly in tasks related to computer vision, such as image classification \cite{SimCLR}, image anomaly detection \cite{Hojjati2022SelfSupervisedAD}, and object recognition \cite{Objectrecognition}. Encouraged by these results, several studies have attempted to apply contrastive learning to the clustering task. An early attempt by Li \etal \cite{CC} showed that contrastive clustering could significantly outperform other baselines on benchmark datasets. Despite these improvements, contrastive clustering and the majority of other deep clustering methods do not consider the interrelationship between data samples, which commonly leads to sub-optimal clustering \cite{DCSS}. However, it would be challenging to incorporate information about the cross-relationship of different instances in an unsupervised setting where we do not have access to data labels. As a result, existing unsupervised CC-based algorithms suffer from high false-negative-pair and low true-positive-pair rates because the model cannot effectively take advantage of the degree of similarity of samples.

\s{INCLUDE and describe here that the existing unsupervised CC-based methods suffer from high false-negative-pair and low positive-pair rate. Explain why this is the case... then in the next paragraph say that our C3 method solves it by discovering the samples similarities ....   apply it to the whole paper in abstract, ...  }

In this paper, we propose a groundbreaking technique to discover similarities between samples. Then, we employ these similarities to identify positive and negative pairs more accurately. We design a soft-weighting scheme that allows focusing on the more challenging to cluster data points. To realize this goal, we create a pool of \textit{weighted} sample pairs where higher weights are assigned to the samples closer to the data cluster boundaries. Such a weighting scheme mitigates the influence of easy-to-cluster data, noise, and, more importantly, the existing false-negative-pair issue in the CC-based methods. Also, by incorporating a larger number of positive samples, we improve the true-positive-pair rate. \s{A  DISCUSSION ON INCREASING TPPR MUST BE ADDED HERE AS WELL. } Overall, our method significantly improves the training efficiency of the model and leads to a cluster-friendly representation. In the remainder of this paper, we describe our idea more formally. Then, we carry out a series of extensive analyses to show that our scheme can significantly improve the clustering performance, and we try to explain how it can achieve such enhancement.

We summarize the contribution of this work as follows: \textbf{(1)} we propose a new contrastive loss function to incorporate the newly discovered positive pairs toward learning a more reliable representation space. \textbf{(2)} we propose a novel weighting scheme that aims to separate more challenging data samples, specifically those that are close to cluster boundaries. This improves the representation learning process and greatly impacts the false-negative-pair selection rate.\s{By doing so, we can improve the effectiveness of sample selection and enhance the quality of the learned representation. Our weighting method also reduces the impact of false negative pairs, noisy samples, and anomalies on the latent representation of the trained network, leading to more robust and accurate results and higher true-positive-rate.} \textbf{(3)} by carrying out extensive experiments, we show that our proposed scheme can significantly outperform current state-of-the-art by a significant margin in terms of several clustering criteria. \s{ inserting almost no extra computations, and}This significant improvement results from considering the pairwise data similarities. \textbf{(4)} We offer insight into the behavior of our developed model, discuss the intuition behind how it improves the clustering performance and support them by conducting relevant experiments.

\s{In this paper, we propose an effective technique to discover similarities between samples and employ them in providing a reliable and less ...  }

\s{\brown{\s{ focus on a suitable subset of challenging negative samples using soft weight assignments, i.e. higher weights are assigned to negative samples that are closer to the boundary of clusters. This weighting scheme also reduces the effect of false negatives, noisy samples, and anomalies on the representation}}. \brown{We} train the algorithm's network so that similar instances that form a cluster become closer together  \brown{and be far from instances of other clusters}.} 

%
\vspace{-6 mm}
\section{Related Works}
\vspace{-2 mm}
\setlength{\parskip}{0pt}
Deep learning-based clustering methods can be categorized into two groups \cite{reviewCluster}: (I) Models that use deep networks for embedding the data into a lower-dimensional representation and apply a traditional clustering such as k-means to the new representation, and (II) Algorithms that jointly train the neural network for extracting features and optimizing the clustering results. In order to achieve a more clustering-friendly representation, previous studies have added regularization terms and constraints to the loss function of neural networks. For example, Huang \etal \cite{DEN} proposed the Deep embedding network (DEN), which imposes a locality preserving and a group sparsity constraint to the latent representation of the autoencoder. These two constraints reduce the inner cluster and increase the inter-cluster distances to improve the clustering performance. In another work, Peng \etal \cite{PARTY} proposed deep subspace clustering with sparsity prior (PARTY) that enhances the clustering efficiency of the autoencoder by incorporating the structure's prior in order to consider the relationship between different samples. Sadeghi and Armanfard \cite{DML} proposed Deep Multi-Representation Learning for Data Clustering (DML), which uses a general autoencoder for instances that are easily clustered along separate AEs for difficult-to-cluster data to improve the performance.
More recent works jointly train the neural network with the clustering objective to further improve the clustering performance. For instance, Deep clustering network (DCN) \cite{DCN} uses k-means objective as the clustering loss and jointly optimizes it with the loss of an autoencoder. Analogously, Deep embedded clustering (DEC) \cite{DEC} first embeds the data into a lower-dimensional space by minimizing the reconstruction loss. Then, it iteratively updates the encoder part of the AE by optimizing a Kullback-Leiber (KL) divergence \cite{KL} loss between the soft assignments and adjusted target distributions. Following the success of DEC, a series of improved algorithms have been developed. For instance, improved deep embedded clustering with local structure preservation (IDEC) \cite{IDEC} jointly optimizes the clustering loss and AE loss to preserve the local structure of data, IDECF \cite{IDECF} adds a fuzzy c-mean network for improving the auxiliary cluster assignment of IDEC during training, and Deep embedded clustering with data augmentation (DEC-DA) \cite{DECDA} applies the DEC method along with the data augmentation strategy to improve the performance. 
Several other methods design auxiliary tasks for learning an efficient representation. E.g., JULE \cite{JULE} applies agglomerative clustering to learn the data representation and cluster assignments. In another algorithm named invariant information clustering (IIC) \cite{IIC}, the mutual information between the cluster assignment of a pair is maximized.
Recently, researchers have turned their attention to self-supervised learning (SSL) models for clustering. For example, MMDC (multi-modal deep clustering) \cite{MMDC} improves the clustering accuracy by solving the proxy task of predicting the rotation. SCAN (semantic clustering by adopting nearest neighbors) \cite{SCAN} first obtains a high-level feature representation using self-supervised learning and then improves the clustering performance by incorporating the nearest neighbor prior.
Contrastive learning is a self-supervised learning paradigm that learns data representation by minimizing the distance between the augmentations of the same sample while pushing them away from other instances. SimCLR \cite{SimCLR} is an example of a contrastive model for learning representation from images that can achieve performance on par with supervised methods. Researchers have increasingly utilized contrastive models for solving tasks such as clustering in the past couple of years. Zhong \etal proposed deep robust clustering (DRC) \cite{DRC} in which a contrastive loss decreases the inter-class variance and another contrastive loss increases the intra-class distance. Contrastive clustering (CC) \cite{CC} improves the clustering performance by jointly performing the instance and cluster-level contrastive learning.
\begin{figure*}[t]
\label{fig11}
  \centering
  \includegraphics[width=0.6\linewidth]{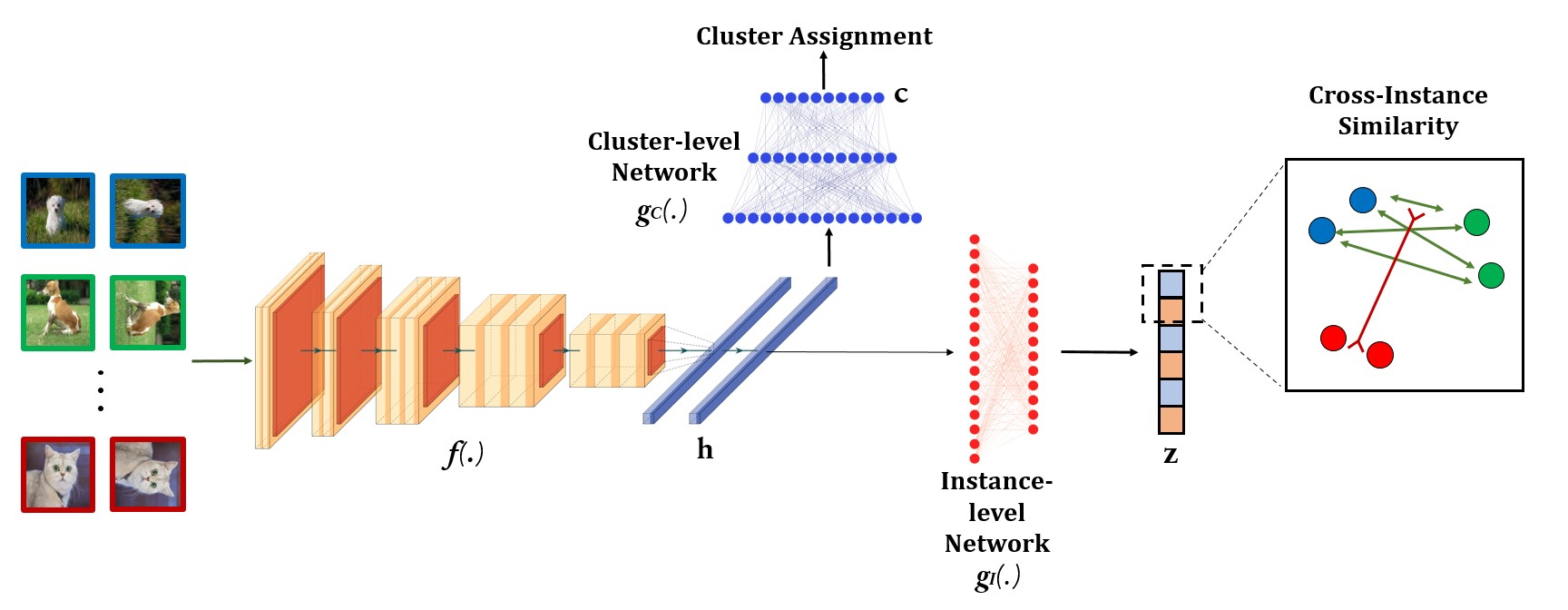}
  \caption{An overview of the training phase of our proposed C3 method.}
  \vspace{-7mm}
\end{figure*}
\vspace{-5mm}
\section{Method}
\label{sec:method}
Given an unlabelled dataset $\mathcal{X} = \{x_1,x_2,\dots,x_N\}$ and a predefined cluster number parameter $\text{M}$, the goal of the clustering problem is to partition $\mathcal{X}$ into $\text{M}$ disjoint groups. Figure \r{2}. shows the overall scheme of the C3 framework.

\s{To realize this goal, our proposed method follows a two-step training procedure:}
%
\s{I suggest, at the begging of the paragraph, first discuss figure 2,  then go over the other matters. In this waa the encoder is defined in the fig2.  }

\s{SHOULD n't T be, $\mathcal{T}$? Have you ever used T throughout the text? $T^a$ and b are not defined}

\s{Notation:} Like other contrastive learning methods, we apply two data augmentations $\mathcal{T}^a$ and $\mathcal{T}^b$, sampled randomly from a pool of transformations, $\mathcal{T}$, to form a \textit{true positive pair} ($x_i^a$, $x_i^b$) for a sample $x_i$, where $x_i^a = \mathcal{T}^a(x_i)$ and $x_i^b = \mathcal{T}^b(x_i)$. In this paper, we used SimCLR transformation pool \cite{SimCLR}. As is shown in Figure \r{2}. , we use the encoder network $f(.)$ to extract features of augmented samples, i.e, $h_i^a = f(x_i^a)$ and $h_i^b = f(x_i^b)$. Inspired by CC \cite{CC}, we devise instance-level and cluster-level contrastive networks, denoted by $g_I(.)$ and $g_C(.)$, respectively. The instance-level network maps the extracted feature of augmented samples to the latent representation (aka z-space), i.e. $z_i^a = g_I(h_i^a)$ and $z_i^b = g_I(h_i^b)$. The output of the cluster-level network is the cluster assignments of samples to different clusters, i.e. $c_i^a=g_C(h_i^a)$ and $c_i^b=g_C(h_i^b)$. We call the output of the cluster-level network c-space. We first initialize our networks, i.e. $f(.)$, $g_I(.)$, and $g_C(.)$ using the CC algorithm.  \s{ MOVE THIS UP:  Hereafter, we call the output of the instance-level network with z-space and the output of the cluster-level network with c-space.}

\s{We first map the data into a latent representation $h$ using an encoder network $f(.)$, so that $h_i = f(x_i; \theta_e) \in \mathbb{R}^d$. Here, $d$ is the dimension of the latent representation, which is usually less than the input size, and $\theta_e$ denotes the parameters of the neural network $f(.)$ and is tuned during the training phase. Ideally, the latent space $h$ should be suitable for clustering while preserving important characteristics of $\mathcal{X}$. To learn such representation, we utilize the self-supervised contrastive clustering (CC) \cite{CC} method. }

\s{Like other contrastive learning methods, CC applies two data augmentations $T^a$ and $T^b$, sampled randomly from a pool of transformations, $\mathcal{T}$, to form a \textit{positive pair} ($x_i^a$, $x_i^b$) for a sample $x_i$, where $x_i^a = T^a(x_i)$ and $x_i^b = T^b(x_i)$. These transformations preserve the important information of the original sample, and human eyes easily perceive their association. However, their pixel values differ significantly, and by learning to create a link between them, the neural network learns to focus on the important and consistent patterns of the data. \s{In this paper, we use the same augmentations as those employed in CC \cite{CC} namely resized crop, gray-scale, horizontal flip, color jittering, and Gaussian blur.}In the next step, the network $f(.)$ extracts representation of augmented samples, i.e, $h_i^a = f(x_i^a)$ and $h_i^b = f(x_i^b)$. After extracting representations, we conduct instance-level and cluster-level contrastive learning: The instance-level contrastive learning could be done by pulling together the representation of positive pairs and pushing them away from negative ones. Since no prior label is given in unsupervised settings, we treat each sample as an individual class, and the two augmentations of the same sample will be considered positive. At the same time, the rest of the batch will be negative. More specifically, for a given sample $x_i$, ($x_i^a$, $x_i^b$) forms its sole positive pair, and the rest of the $2N_1-2$ samples will be considered negative. In practice \cite{SimCLR}, instead of directly applying the contrastive learning on the representation $h$, we first map it to another subspace using a multi-layer perceptron $g_I$ to obtain $z_i = g_I(h_i)$ and then minimize the following loss function:}

\s{
\begin{equation}
\label{ccloss}
    \ell_{i}^{a} = -\log \frac{\exp(\mathrm{sim}(z_i^a,z_i^b)/\tau_I)}{\sum_{k \in \{a,b\}} \sum_{j=1}^{N}\exp(\mathrm{sim}(z_i^a,z_j^k) / \tau_I)},
\end{equation}
\begin{equation}
    \mathrm{sim}(z_i^{k_1},z_j^{k_2}) = \frac{(z_i^{k_1})^\intercal(z_j^{k_2})}{\|z_j^{k_1}\| \|z_i^{k_2}\|},
\end{equation}
}
\s{where $\tau_I$ is the temperature parameter that defines the degree of attraction and repulsion between samples. If we normalize $z_i$s, the similarity metric will become $\mathrm{sim}(z_i^{k_1},z_j^{k_2})=(z_i^{k_1})^\intercal(z_j^{k_2})$.}

\s{The final loss function is defined as the average of the loss for all positive pairs across the batch:
}
\s{\begin{equation}
    \mathcal{L}_{ins} = \frac{1}{2N} \sum_{i=1}^{N}(\ell_i^a + \ell_i^b)
\end{equation}
}
\s{The reason that $z_i$ is used instead of $h_i$ in Eq. (\ref{ccloss}), is because minimizing the contrastive loss on the representation $h_i$ might cause it to drop essential information. Previous studies have also empirically shown that minimizing $z_i$ leads to better results \cite{SimCLR}.
}

\s{Decoupled from the instance-level contrastive learning network $g_I(.)$, we train another network $g_C(.)$ that maps the h-space onto an M-dimensional c-space. Mathematically speaking, if we denote the output of $g_C$ under the first augmentation as $Y^a$, then $Y^a \in \mathcal{R}^{N \times M}$. Intuitively, in this subspace, the $i$-th element corresponds to the probability of the sample belonging to the $i$-th cluster.}

\s{If we show the $i$-th column of $Y^a$ by $y_i^a$, we can consider the representation of the second augmentation $y_i^b$ as its positive pair while leaving the other $2M-2$ columns as negative. Then, similar to the instance-level loss, we define a contrastive loss function to pull the positive pair and push it away from the negatives as follows:}

\s{\begin{equation}
    \hat{\ell}_{i}^{a} = -\log \frac{\exp(\mathrm{sim}(y_i^a,y_i^b)/\tau_C)}{\sum_{k \in \{a,b\}} \sum_{j=1}^{N}\exp(\mathrm{sim}(y_i^a,y_j^k) / \tau_C)}
\end{equation}}

\s{Minimizing the above loss can lead to the trivial solution of mapping most data points to the same cluster. To avoid this, CC minimizes the negative entropy of cluster assignment probabilities defined below in Eq. (\ref{entropy}) where $P(y_i^k) = \frac{1}{N}\sum_{j=1}^N Y_{ji}^k$.}
\s{\begin{equation} \label{entropy}
    H(Y) = -\sum_{i=1}^M [P(y_i^a)\log P(y_i^a) + P(y_i^b)\log P(y_i^b)].
\end{equation}}

\s{The final cluster-level contrastive loss is calculated as:}

\s{
\begin{equation}
    \mathcal{L}_{clu} = \frac{1}{2M}\sum_{i=1}^M (\hat{\ell}_i^a+\hat{\ell}_i^b) - H(Y).
\end{equation}}

\s{The final loss of CC is the sum of the instance-level and cluster-level loss:}
\s{\begin{equation}
    \mathcal{L}_{CC} = \mathcal{L}_{ins} + \mathcal{L}_{clu}
\end{equation}}

\s{As the first step in our method, we initialize our networks by training them to minimize the CC loss. }If we do such initialization process for a sufficient number of epochs, a partially reliable z-space will be obtained. However, the z-space obtained by minimizing the CC loss is sub-optimal for clustering, which results in a large false-negative-pair rate and a low true-positive-pair rate. To mitigate this issue, our method incorporates cross-sample similarities.  \s{REFRE TO THE ISSUE of FPPR and FNPR and that we use cross sample similarities to mitigate the issue and hen the next paragraph that it is unsupervised so we have to use a notion of self-supervision in refining positive and negative pairs... .... the cross-sample relationships in its training phase.}

Since our framework is unsupervised and we do not have access to the data labels to identify samples belonging to the same cluster, we use a notion of self-supervision to refine clusters. We employ the cross-sample similarities in the partially trained z-space to realize the self-supervision concept. The cross-sample similarity is measured by the cosine distance of samples in the z-space. If, for a pair of instances, the similarity is greater than or equal to a threshold $\zeta$, we consider those samples to be similar and pull them closer together by minimizing the loss function $\mathcal{L}_{C3}$ defined below:
{\small
\begin{equation}
\label{totalloss}
    \mathcal{L}_{C3} = \frac{1}{2N}\sum_{i=1}^N (\Tilde{\ell}_i^a+\Tilde{\ell}_i^b)
\end{equation}}
\s{\begin{equation}
    \label{C3loss}
    \Tilde{\ell}_{i}^{a} = -\log \frac{\sum_{k \in \{a,b\}} \sum_{j=1}^{N}\mathrm{\mathds{1}\{z_i^{a^\intercal} z_j^k \geq \zeta\} \exp(z_i^{a^\intercal} z_j^k)}}{\sum_{k \in \{a,b\}} \sum_{j=1}^{N}\mathrm{\exp(z_i^{a^\intercal} z_j^k)}},
\end{equation}}
{\small
\begin{equation}
    \label{C3loss}
    \Tilde{\ell}_{i}^{a} = -\log \frac{\sum_{k \in \{a,b\}} \sum_{j=1}^{N}\mathrm{\mathds{1}\{z_i^{a^\intercal} z_j^k \geq \zeta\} \exp(z_i^{a^\intercal} z_j^k)}}{\sum_{k \in \{a,b\}} \sum_{j=1}^{N}\mathrm{w_{ij}^k\exp(z_i^{a^\intercal} z_j^k)}},
\end{equation}}
where $\mathds{1}\{.\}$ denotes the indicator function.\s{Note that since $z_i^k$s are normalized, $\mathrm{sim}(z_i^a,z_j^k)=(z_i^a)^\intercal(z_j^k)$.} Analogous to $\Tilde{\ell}_{i}^{a}$, we define $\Tilde{\ell}_{i}^{b}$ that considers similarity of $z_i^b$ and other samples in the batch.
Furthermore, in the denominator of the proposed loss function, we included $w_{ij}^k$ to consider higher weights for the samples that are neither close together nor far from each other. In this way, we realize the goal of decreasing the false negative pair selection rate as, in the traditional CC-based methods, all augmented samples in the batch are equally considered when forming negative pairs, regardless of the possibility of them belonging to the same cluster and the difficulty of the samples to cluster.

We assume that the weight terms are given when minimizing the C3 loss defined in \eqref{C3loss}. To obtain an optimum value for a weight term $w_{ij}^k$, we propose solving the below optimization problem while the networks are frozen. The first term of the below optimization problem is defined based on our motivation to assign a very low weight to too close and too far away samples and, instead, let the remaining samples, which are more probably located on the cluster boundaries, take higher weights. The second term is to avoid the trivial solution of assigning a weight equal to one to the sample providing the maximum value of $1-|z_i^{a^\intercal} z_j^k|$. To avoid instability, we include the constraint by which the summation of all weights must be equal to 1. 
\vspace{-3mm}
{\small
\begin{align}
   \min_{w_{ij}^k} \sum_{k\in \{a,b\}}\sum_{j=1}^{N} -w_{ij}^k(&1-|z_i^{a^\intercal} z_j^k|)-\frac{1}{\Gamma}\,\, 
 \text{H}(W_i) & s.t. \sum_{k\in \{a,b\}}\sum_{j=1}^{N} w_{ij}^k = 1 \label{eq11}
\end{align}}
\s{as a weighting term that assigns weight to \o{the sample pairs}\s{WRONG: the negative samples -- WE MUST NOT CALL THEM NEGATIVE PAIRS AS IT INCLUDES POSITIVE PAIRS AS WELL.} based on their importance in determining the boundaries of the separation. \brown{For finding $w_{ij}^k$, we detach z-space from the gradient graph of our networks; hence, we do not update the parameters of our model with respect to $w_{ij}^k$.} In this weighting scheme, the network should not repel the samples that are very close and are deemed \o{to be} positive, i.e. $\,z_i^{a^\intercal} z_j^k \xrightarrow{}1$, and, on the other hand, should not focus on maximizing the distance between the samples that are already mapped to \s{a further ??? OR farther ??? } \o{two far points in the z-space}\s{  representation}, i.e. $\,z_i^{a^\intercal} z_j^k \xrightarrow{}-1$. Instead, it should focus on the points that are near the boundary. For finding $w_{ij}^k$, we \o{suggest} to solve the following optimization problem that aims at \o{mitigating} \s{relaxing ?? IT IS NOT A GOOD TERM} the effect of\s{false ?????? ARE U Sure? **} false-negative-pair selection and \o{down-weighting} samples that are \o{already} mapped to far locations in the z-space.\s{DISCUSS a bit that WHY this optimization problem?? e.g. For finding $w_{ij}^k$, we need to solve an optimization problem that is not unique and aims at reducing the impact of false negative samples, i.e  
, and learning samples  making the **** avoid collapsing issue an, ...  then say that we suggest below while not unique...}}
In the above equation, $\Gamma$ is a hyperparameter, $W_i = \{w_{ij}^k|j\in\{1,2,..,N\}, k\in (a,b)\}$ is the set of all weights, and $\text{H}(.)$ is the entropy function.\s{ that avoids converging to a trivial solution \b{where one weight that corresponds to the maximum value of $1-|z_i^{a^\intercal} z_j^k|$ converges to one and other weights converges to zero}. The first term aims to assign lower weights to similar samples (i.e. $z_i^{a^\intercal} z_j^k \xrightarrow{}1$) and samples that are far enough from each other (i.e. $z_i^{a^\intercal} z_j^k \xrightarrow{}-1$)}\s{ REVISE the following based on the new writings we discussed on low TPR and high FNR: By solving this optimization problem, we relax the effect of false-negative-pairs, noisy samples, and anomalies in the learned representation.}
We solve the optimization problem defined in \eqref{eq11} using the Lagrange multiplier technique as below:
%
{\small
\begin{align}
    L = \sum_{k\in \{a,b\}}\sum_{j=1}^{N} -w_{ij}^k(1-|z_i^{a^\intercal} z_j^k|)+\frac{1}{\Gamma}\sum_{k\in \{a,b\}}\sum_{j=1}^{N}w_{ij}^k\log(w_{ij}^k) +\lambda(\sum_{k\in \{a,b\}}\sum_{j=1}^{N} w_{ij}^k - 1) 
\end{align}}
\vspace{-5mm}
{\small \begin{align} \nonumber
\frac{\partial L}{\partial w_{ij}^k} = 0\xrightarrow[]{} & -(1-|z_i^{a^\intercal} z_j^k|) +\frac{1}{\Gamma}\log(w_{ij}^k) +\frac{1}{\Gamma} +\lambda =0\\  & w_{ij}^k = \exp(\Gamma(1-|z_i^{a^\intercal} z_j^k|)-1-\Gamma\lambda) \label{eq13}
\end{align}}
Where $\lambda$ is the Lagrange multiplier. By substituting \eqref{eq13} into the constraint of \eqref{eq11}, we have:
{\small \begin{align}\nonumber
       &\sum_{k\in \{a,b\}}\sum_{j=1}^{N} w_{ij}^k = 1 \xrightarrow[]{} \sum_{k\in \{a,b\}}\sum_{j=1}^{N} \exp(\Gamma(1-|z_i^{a^\intercal} z_j^k|)-1-\Gamma\lambda) = 1 \nonumber\\
       &\exp(-1-\Gamma\lambda) = \frac{1}{\sum_{k\in \{a,b\}}\sum_{j=1}^{N} \exp(\Gamma(1-|z_i^{a^\intercal} z_j^k|))} \label{eq14}
\end{align}}
If we substitute \eqref{eq14} to \eqref{eq13}, we have the final values for $w_{ij}^k$ as below:
{\small \begin{align} \label{eq:weighting}
    w_{ij}^k = \frac{\exp(\Gamma(1-|z_i^{a^\intercal} z_j^k|))}{\sum_{k\in \{a,b\}}\sum_{j=1}^{N} \exp(\Gamma(1-|z_i^{a^\intercal} z_j^k|))}
\end{align}}

\s{The reason we use the z-space for similarity measurements rather than the h-space is mainly that computing the inner products in the z-space requires fewer mathematical operations as the dimensions of z are lower than that of h.}
\s{REVISE this paragrapgh TO ACCURATELY INCLUDE WHAT WE ACTually have ...}When comparing the loss function of other contrastive clustering algorithms, such as CC, with the loss of C3 (shown in Eq. \eqref{C3loss}), several differences become apparent. Firstly, while other contrastive algorithms typically only allow one positive pair to appear in the numerator, C3 enables the networks to be trained by considering a much larger number of positive pairs. This leads to more efficient training and better performance. Secondly, C3 adopts a weighting scheme when creating the negative pairs; the scheme assigns lower weights to samples that are either too close or too far from each other while assigning higher weights to those that are more in the mixing cluster areas\s{located on the boundaries of clusters}. This approach reduces the impact of false positive, noisy, and anomaly samples on the learning of representations. \s{*** Isn't this following paragraph redundant with the first and second items above?** remove if agree yes.*** Additionally, C3 \b{improves the clustering accuracy by simultaneously focusing on the negative samples that are located near the cluster boundaries (effect of the denominator) and increasing the number of true-positive pairs (effect of the numerator).}}
\s{The numerator Eq. (\ref{C3loss}) confirms that C3 allows the networks to get trained considering much more positive pairs, while other contrastive algorithms commonly allow only one positive pair to appear in the numerator. This is indeed a valuable property of C3, as considering all batch samples as negative samples is a misleading assumption due to the fact that samples of one cluster should indeed be considered as positive pairs and pulled together. It is worth noting that, from the number of required operations point of view, computing the C3 loss needs no extra computations compared to the vanilla contrastive loss -- C3 only needs to find positive pairs and perform a few more summations, in the numerator, equal to the number of extra positive pairs. }

\s{\begin{algorithm}[t]

\KwIn{Hyperparameter $\zeta$, dataset $\mathcal{X}$, transformation pool $\mathcal{T}$, initializing epochs $E_{1}$, training epochs $E_{2}$, batch size $N_1$ and $N_2$, learning rate $\eta_1$ and $\eta_2$, temperature parameters $\tau_I, \tau_C$, number of clusters $\text{M}$, $f,g_I$ and $g_C$.}
\KwOut{Cluster assignment $c$}
\caption{\r{C3 Pseudo code}}
\label{pcode}
\tcp{Training}
Apply contrastive clustering (CC) algorithm for $E_1$ epochs to train $f,g_I,g_C$\\
\For{epoch $=1$ \KwTo $E_{2}$}
{
Sample a mini-batch $\{x_i\}_{i=1}^{N_2}$ from $\mathcal{X}$\\
Sample two augmentations $T^a, T^b \sim \mathcal{T}$\\
Apply $T^a, T^b$ to $x_i$ to get:\\
\Indp $x_i^a = T^a(x_i), x_i^b = T^b(x_i)$\\
\Indm
Compute the instance representation by:\\
\Indp $h_i^a = f(T^a(x_i)), h_i^b = f(T^b(x_i))$ \\
$z_i^a = g_I(h_i^a), z_i^b = g_I(h_i^b)$ \\
\Indm
Compute the loss $\Tilde{\ell}_i^a$ and $\Tilde{\ell}_i^b$ through Eq. \ref{C3loss}\\
Compute the overall loss $\mathcal{L}_{C3}$ using Eq. \ref{totalloss}\\
Update $f, g_I$ to minimize $\mathcal{L}_{C3}$
}
\tcp{Test}
\For{$x$ in $\mathcal{X}$}
{
Calculate the latent representation by\\
\Indp $h=f(x)$\\
\Indm
Find the cluster assignment by\\
\Indp $c = \arg \max g_C(h)$
}
\end{algorithm}}

\s{In summary, in our framework, we initialize the networks by CC. This is mainly to train $g_C(.)$ and obtain a partially reliable z-space that will be used in the following step when creating more positive pairs to be employed in the C3 loss. In the second step, we boost the z-space ability in providing reliable representations by identifying similar samples and then bringing them closer together to form clusters with more explicit boundaries. Pseudocode of the proposed C3 method is presented in Algorithm \ref{pcode}.}
\vspace{-3mm}
\section{Experiments and Discussions}
\vspace{-3 mm}
\label{sec:experiments}
In this section, we demonstrate the effectiveness of our proposed scheme by conducting rigorous experimental evaluations.
We evaluated our method on five challenging computer vision benchmark datasets: CIFAR-10, CIFAR-100 \cite{Cifar}, ImageNet-10, ImageNet-Dog \cite{Imagenet}, and Tiny-ImageNet \cite{Tinyimage}. 
For CIFAR-10 and CIFAR-100, we combined the training and test splits. Also, for CIFAR-100, instead of 100 classes, we used the 20 super-classes as the ground-truth. To evaluate the performance, we use three commonly-used metrics in clustering namely clustering accuracy (ACC), Normalized Mutual Information (NMI), and Adjusted Rand Index (ARI) \cite{reviewCluster}. 
\s{\subsection{Implementation Details}}

For the sake of fair comparison, for all datasets, we used ResNet34 \cite{resnet} as the backbone of our encoder $f(.)$, which is the same architecture that previous algorithms have adopted. We set the dimension of the output of the instance-level projection head $g_I$ to 128, for all datasets. The output dimension of the cluster-level contrastive head is set to the number of classes, M, in each dataset. All networks are initialized by the CC \cite{CC} algorithm with the hyperparameters suggested by its authors. The Adam optimizer with an initial learning rate of  $0.00001$ and batch size of $128$ is used for C3. All networks are trained for 20 epochs.\s{Like CC, we set the temperature parameters of instance-level and clustering-level networks to $\tau_I = 0.5$ and $\tau_C = 1.0$, respectively. Like CC, for training the first step, we used Adam optimizer \cite{kingma:adam} with $\eta_{1} = 0.0003$ , batch size of $N_{1}=256$ and $E_{1} = 1000$ epochs. Also, the Adam optimizer with an initial learning rate of $\eta_{2} = 0.00001$ and batch size of $N_{2} = 128$ is used in the second step, and the networks are trained for $E_{2} = 20$ epochs.} The experiments are run on NVIDIA TESLA V100 32G GPU.
\vspace{-4mm}
\subsection{Comparison with State-of-the-art}
\label{sec:sota}
\begin{table*}[!t]
\caption{Clustering performance of different methods.}
\vspace{-2mm}
\label{tab:results}
\setlength{\tabcolsep}{0.8pt}
\begin{center}
{\scriptsize
\makebox[\textwidth][c]{
\begin{tabular}{c|ccc|ccc|ccc|ccc|ccc|cc}
    \toprule 
      &\multicolumn{3}{c|}{CIFAR-10} & \multicolumn{3}{c|}{CIFAR-100} & \multicolumn{3}{c|}{ImageNet-10}& \multicolumn{3}{c|}{ImageNet-Dogs}& \multicolumn{3}{c|}{Tiny-ImageNet}\\
     \cline{2-18}
    Algorithm & NMI  & ACC & ARI & NMI  & ACC & ARI &NMI  & ACC & ARI &NMI  & ACC & ARI &NMI  & ACC & ARI \\
    \midrule
    K-means \cite{kmeans}&0.087&0.229&0.049&0.084&0.130&0.028&0.119&0.241&0.057&0.055&0.105&0.020&0.065&0.025&0.005\\
    SC \cite{SC}&0.103&0.247&0.085&0.090&0.136&0.022&0.151&0.274&0.076&0.038&0.111&0.013&0.063&0.022&0.004\\
    AC \cite{AC}&0.105&0.228&0.065&0.098&0.138&0.034&0.138&0.242&0.067&0.037&0.139&0.021&0.069&0.027&0.005\\
    NMF \cite{nmf}&0.081&0.190&0.034&0.079&0.118&0.026&0.132&0.230&0.065&0.044&0.118&0.016&0.072&0.029&0.005\\
    AE \cite{ae}&0.239&0.314&0.169&0.100&0.165&0.048&0.210&0.317&0.152&0.104&0.185&0.073&0.131&0.041&0.007\\
    DAE \cite{dae}&0.251&0.297&0.163&0.111&0.151&0.046&0.206&0.304&0.138&0.104&0.190&0.078&0.127&0.039&0.007\\
    DCGAN \cite{dcgan}&0.265&0.315&0.176&0.120&0.151&0.045&0.225&0.346&0.157&0.121&0.174&0.078&0.135&0.041&0.007\\
    DeCNN \cite{decnn}&0.240&0.282&0.174&0.092&0.133&0.038&0.186&0.313&0.142&0.098&0.175&0.073&0.111&0.035&0.006\\
    VAE \cite{vae}&0.254&0.291&0.167&0.108&0.152&0.040&0.193&0.334&0.168&0.107&0.179&0.079&0.113&0.036&0.006\\
    JULE \cite{JULE}&0.192&0.272&0.138&0.103&0.137&0.033&0.175&0.300&0.138&0.054&0.138&0.028&0.102&0.033&0.006\\
    DEC \cite{DEC}&0.275&0.301&0.161&0.136&0.185&0.050&0.282&0.381&0.203&0.122&0.195&0.079&0.115&0.037&0.007\\
    DAC \cite{dac}&0.396&0.522&0.306&0.185&0.238&0.088&0.394&0.527&0.302&0.219&0.275&0.111&0.190&0.066&0.017\\
    ADC \cite{adc}&-&0.325&-&-&0.160&-&-&-&-&-&-&-&-&-&-\\
    DDC \cite{DDC}&0.424&0.524&0.329&-&-&-&0.433&0.577&0.345&-&-&-&-&-&-\\
    DCCM \cite{DCCM}&0.496&0.623&0.408&0.285&0.327&0.173&0.608&0.710&0.555&0.321&0.038&0.182&0.224&0.108&0.038\\
    IIC \cite{IIC}&-&0.617&-&-&0.257&-&-&-&-&-&-&-&-&-&-\\
    PICA \cite{PICA}&0.591&0.696&0.512&0.310&0.337&0.171&0.802&0.870&0.761&0.352&0.352&0.201&0.277&0.098&0.040\\
    GATCluster \cite{gatcluster2020}&0.475&0.610&0.402&0.215&0.281&0.116&0.609&0.762&0.572&0.322&0.333&0.200&-&-&-\\
    CC \cite{CC} &0.678*&0.770*&0.607*&0.421*&0.423*&0.261*&0.850*&0.893*&0.811*&0.436*&0.421*&0.268*&0.331*&0.137*&0.062*\\
    EDESC \cite{edesc}&0.627&0.464&-&0.385&0.370&-&-&-&-&-&-&-&-&-&-\\
    \hline
    C3 (Ours) &\textbf{0.743}&\textbf{0.836}&\textbf{0.703}&\textbf{0.435}&\textbf{0.456}&\textbf{0.274}&\textbf{0.905}&\textbf{0.943}&\textbf{0.860}&\textbf{0.447}&\textbf{0.434}&\textbf{0.280}&\textbf{0.335}&\textbf{0.140}&\textbf{0.064}\\
    \bottomrule
\end{tabular}
}}
\end{center}
\end{table*}

\begin{figure*}[t]
\vspace{-6mm}
  \centering
  \includegraphics[width=0.8\linewidth]{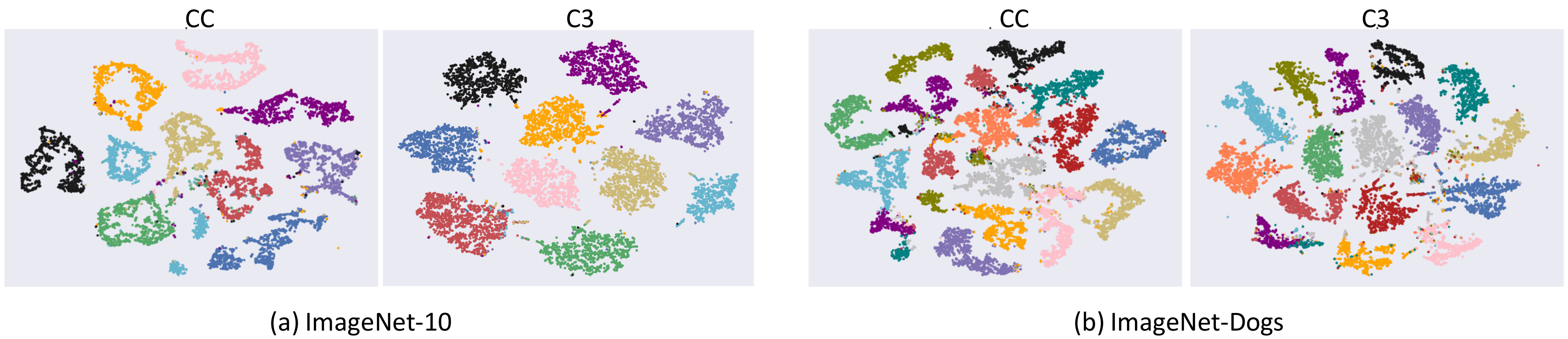}
  \caption{t-SNE visualization of clusters learned by the CC and C3 methods.}
  \vspace{-5mm}
  \label{fig:tsne}
\end{figure*}

\begin{wrapfigure}{R}{0.5\textwidth}
\vspace{-4mm}
\begin{minipage}[b]{.48\linewidth}
  \centering
  \centerline{\includegraphics[width=3.0cm]{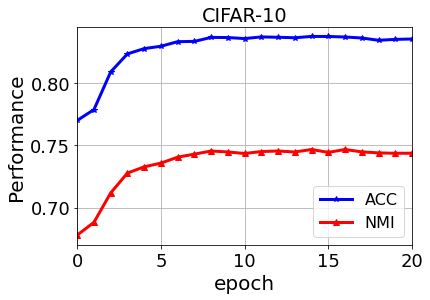}}
\end{minipage}
\hfill
\begin{minipage}[b]{0.48\linewidth}
  \centering
  \centerline{\includegraphics[width=3.0cm]{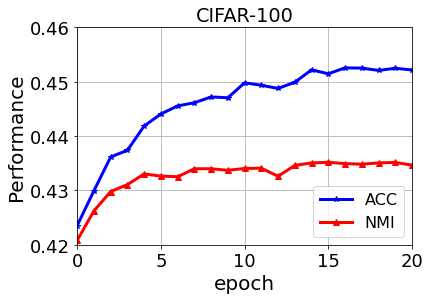}}
\end{minipage}
\begin{minipage}[b]{.48\linewidth}
  \centering
  \centerline{\includegraphics[width=3.0cm]{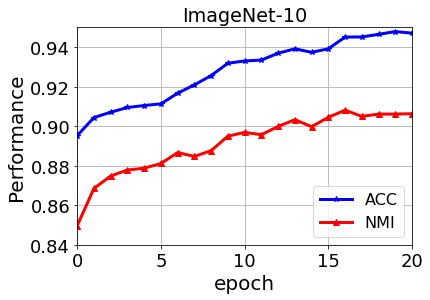}}
\end{minipage}
\hfill
\begin{minipage}[b]{0.48\linewidth}
  \centering
  \centerline{\includegraphics[width=3.0 cm]{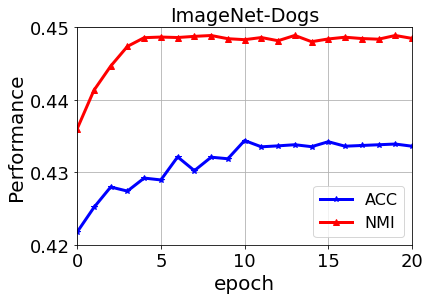}}
\end{minipage}
\caption{C3's performance vs epochs.}
\label{fig:convergence}
\end{wrapfigure}

Table \ref{tab:results} shows the results of our proposed method on benchmark datasets, compared to state-of-the-art and some common traditional clustering methods. For CC, we run the code provided by its authors for all datasets, and the results are indicated by (*). As is evident in this table, our proposed method significantly outperforms all other baselines in all datasets. Quantitatively, comparing to the second best algorithm (i.e. CC), C3 improves the ACC by \s{ARE THE FOLLOWING NUMBERS UPDATED BASED ON THE CURRENT TABLE?? }$6.6\%$, $3.3\%$, $5.0\%$, $1.3\%$ and $0.3\%$), the NMI by $6.5\%$, $1.4\%$, $5.5\%$, $1.1\%$ and $0.4\%$, and the ARI by $9.6\%$, $1.3\%$, $4.9\%$, $1.2\%$ and $0.2\%$, respectively on CIFAR-10, CIFAR-100, ImageNet-10, ImageNet-Dogs, and Tiny-ImageNet. The main difference between our framework and other baselines, such as CC, is that we exploit an additional set of information, i.e. the similarity between samples, to further enhance the learned representation for clustering. In our approach, we increase the number of positive samples, which improves the true-positive-pair rate, and put more weight on clustering the samples that are more probably located on the boundary of the clusters, which leads to a decrease in the false-negative-pair selection rate. We believe this is the main reason our method's performance is superior compared to the baselines.

As opposed to the CC method that misses the global patterns present in each data cluster, C3 correctly tries to consider such patterns (at least partially) by employing cross-instance data similarities.\s{and assigning lower weights to false negative samples and samples that are located far enough from each other in the z-space.}\s{ AND ****FILL***.} Looking at \eqref{C3loss} and \eqref{eq11}, one can infer that C3 implicitly reduces the intra-cluster distance while maximizing the inter-cluster distance, which is what an efficient grouping technique would do in the presence of the data labels. This can be confirmed by visualizing the clusters before and after applying the C3 loss. As Figures \ref{fig:tsne_cifar} and \ref{fig:tsne} show, the samples are fairly clustered after initialization with CC\s{at the end of the first step training using the CC loss,}, i.e. before the start of training using the C3 loss. However, some of the difficult clusters are mixed in the boundaries. After initialization, clusters \s{obtained by minimizing only the CC loss} are expanded with a considerable number of miss-clustered data. However, after training with the proposed C3 method, we observe that the new cluster space is much more reliable, and individual clusters are densely populated while being distant from each other.

\subsection{Convergence Analysis}

Results of section \ref{sec:sota} depict the superiority of our proposed scheme. Now, we analyze C3's convergence and the computational complexity to evaluate at what cost it makes such an improvement over other baselines. We plotted the trend of clustering accuracy and NMI for four datasets during the training epochs in Figure \ref{fig:convergence}. We can readily confirm that although we are just training the C3 step for $20$ epochs, the graphs quickly converge to a settling point, which corresponds to the peak performance. Also, we can observe that both ACC and NMI are improved throughout the C3 training phase in all datasets. The performance at $\mathrm{epoch}=0$ corresponds to the clustering performance after initialization with CC\s{final performance of the CC algorithm}. These figures clearly show that C3 improves its clustering quality and justifies the qualitative results shown in Section \ref{sec:sota}.

\begin{wrapfigure}[]{}{0.5\textwidth}
\vspace{-9mm}
    \centering
    \includegraphics[scale=0.25]{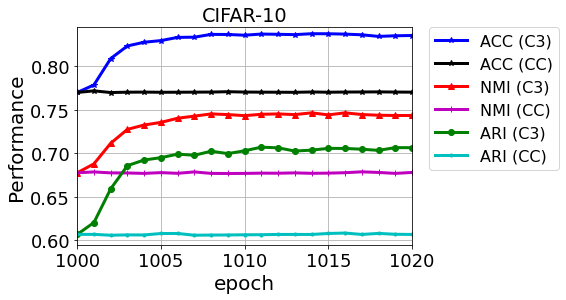}
    \vspace{-5mm}
    \caption{Performance of CC and C3 after 1020 epochs.}
    \label{fig:vscc}
\end{wrapfigure}

One may naively \s{(wrongly)}think that the better performance of C3 \s{compared to CC} is because it is being trained for 20 more epochs; note that in all our experiments, as suggested by the CC authors,  we trained the CC algorithm networks for 1000 epochs. We train the C3 networks for 1020 epochs.\s{i.e. 1000 epochs for the first step and 20 epochs for the second step.} We reject this argument and support it by training the CC networks for the same number of epochs as what the C3 is trained for, i.e. 1020 epochs. We observe that no improvement is obtained for CC when trained for an extra 20 epochs. \s{This is while when we keep training the networks with the C3 loss for only 20 more epochs, significant improvements are observed. This shows that the superior performance of C3 is not because of the extra $2\%$ epochs but because its objective function helps the network discover patterns that are complementary to those that CC extracts.} The result of such an experiment on CIFAR-10 is shown in Figure \ref{fig:vscc}.

\subsection{How does C3 loss improve the clusters?}
\label{sec:improve}



As we saw in Figure \ref{fig:tsne}, the improvement that C3 achieves is mainly because it is able to reduce the distance between instances of the same cluster while repelling them from other clusters. We can justify this observation by considering the loss function of C3, i.e. Eq. (\ref{C3loss}). In this function, the term $\mathds{1}\{z_i^{a^\intercal} z_j^k \geq \zeta\}$ indicates that if the cosine similarity of two samples is greater than $\zeta$, they should be moved further close to each other.
\begin{wrapfigure}[9]{}{0.5\textwidth}
  \centering
  \vspace{-6mm}
  \includegraphics[scale=0.25]{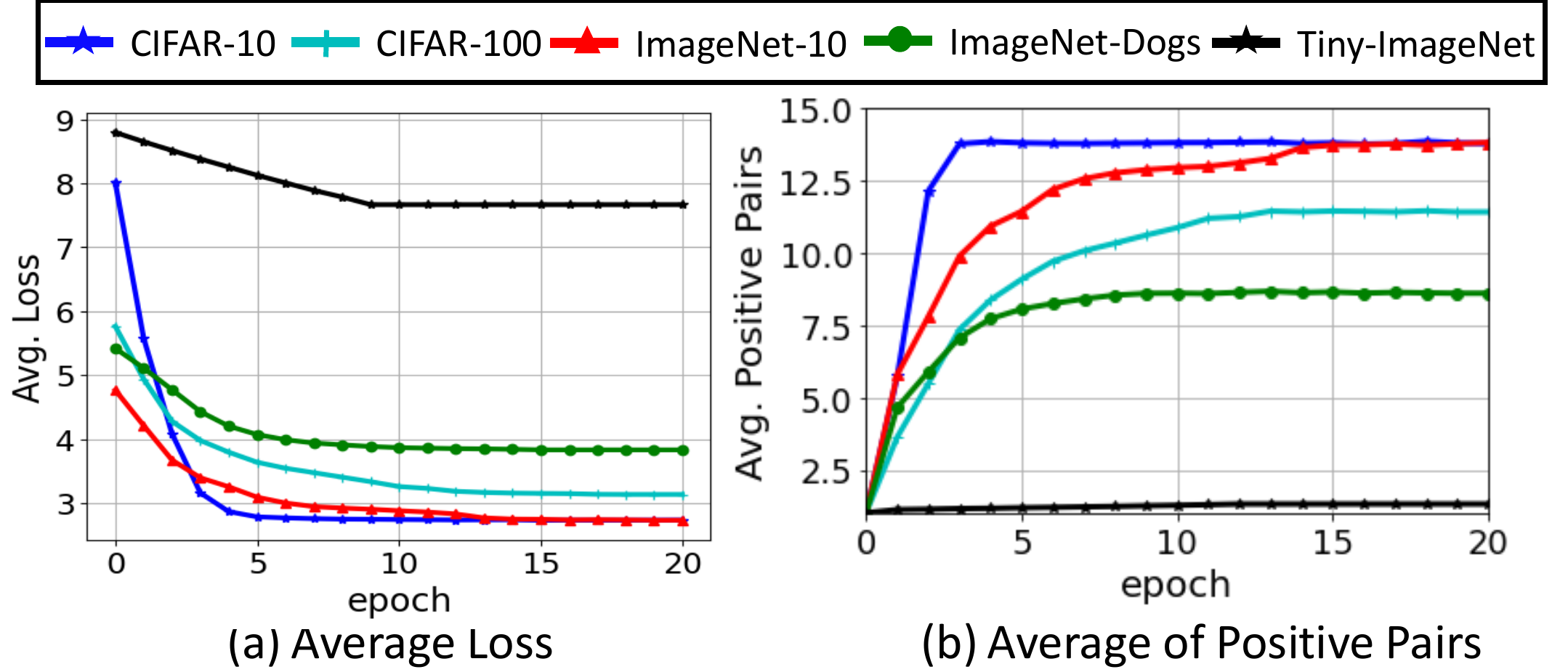}
  \caption{Plot of loss and number of positive pairs versus epoch.}
  \vspace{-4mm}
  \label{fig:pairs}
\end{wrapfigure}
At the beginning of training with the C3 loss, since the networks are initiated with CC,\s{, since the z-space is initiated with the \r{THIS PARAGRAPH NEEDS REVISION}CC obtained z-space,} we can assume that the points that are very similar to each other have a cosine distance, in the z-space, less than threshold $\zeta$, so they will become further close by minimizing the C3 loss. For instance, take two points $z_1$ and $z_2$ with a cosine similarity larger than $\zeta$, and assume that $z_3$ has a similarity greater than $\zeta$ with $z_1$, but its similarity with $z_2$ is smaller than $\zeta$. Therefore, according to the loss function, $z_1$ and $z_2$, as well as $z_1$ and $z_3$, are forced to become closer, but it is not the case for $z_2$ and $z_3$. However, these two points will also implicitly move closer to each other because their distance to $z_1$ is reduced. As the training continues, at some point, the similarity of $z_2$ and $z_3$ also may pass the threshold $\zeta$. Therefore, as the similar pairs move closer to each other during the training, a series of new connections will be formed, and the cluster will become denser. To support this hypothesis, we plotted the average of the loss function and the average number of positive pairs of each data sample in Figure \ref{fig:pairs}-a and \ref{fig:pairs}-b, respectively. We can observe that the number of positive pairs exponentially increases during the training until it settles to form the final clusters. Corresponding to this exponential increase, we can see that the loss is decreasing, and the network learns a representation in which clusters are distanced from each other while samples of each cluster are packed together.

We can also deduct from this experiment that the number of positive pairs is also related to the number of classes in each dataset. For example, if we have an augmented batch size of $N=256$, for Tiny-ImageNet that has 200 classes, we expect to have $\frac{256}{200} = 1.28$ positive pairs per sample which is very close to $1$ and it is the reason that we do not see the same sharp increasing trend as other datasets in Tiny-ImageNet.
\vspace{-5mm}
\subsection{Effect of Hyperparameter $\zeta$ and $\Gamma$}
\vspace{-2mm}

\begin{figure*}[t!]
  \centering
  \includegraphics[scale=0.18]{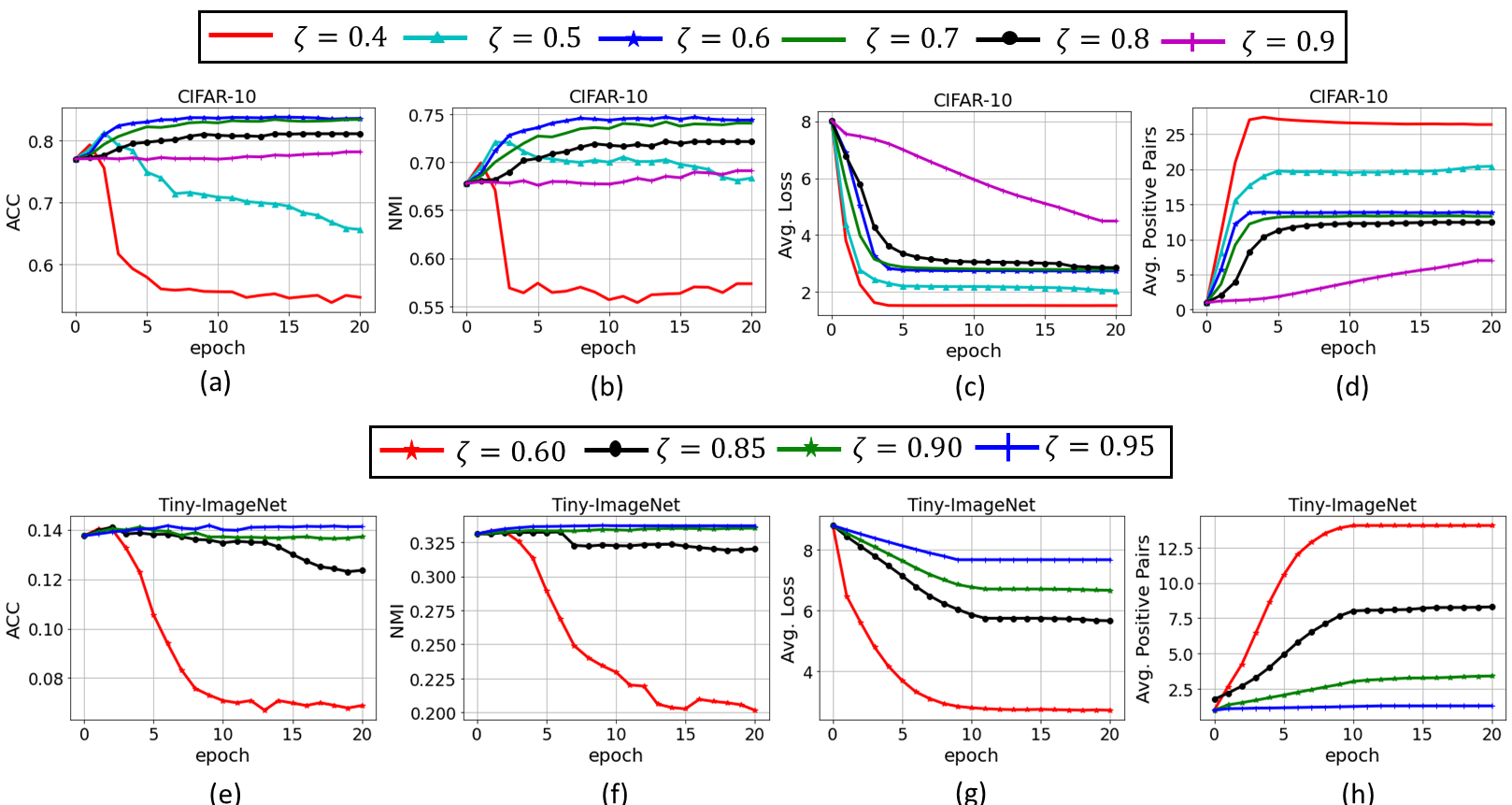}
  \vspace{-2mm}
  \caption{Performance and behavior of C3 for different values of $\zeta$, for CIFAR 10 (top row) and Tiny-ImageNet (bottom row).} \vspace{-6mm}
  \label{fig:zeta}
\end{figure*}

Our method, C3, introduces two new hyperparameter $\zeta$ and $\Gamma$. $\zeta$ is a threshold for identifying similar samples. Throughout the experiments, we fixed $\zeta=0.6$, which yielded consistent results across datasets. Now, we carry out an experiment in which we change $\zeta$ and record the performance. Note that since $z_i^{a^\intercal} z_j^k \in [-1,1]$, we can technically change $\zeta$ from -1 to 1. Intuitively, for a small or negative value of $\zeta$, most points in the z-space will be considered similar, and the resulting clusters will not be reliable. Therefore, in our experiment, we change $\zeta$ from 0.4 to 0.9 in 0.1 increments for CIFAR-10. For Tiny-ImageNet, as there are lots of clusters, we set $\zeta \in \{0.60, 0.85, 0.90, 0.95\}$. We then plot the accuracy, NMI, average loss, and the average of positive pairs per sample. The graphics are shown in Figure \ref{fig:zeta}.

In CIFAR-10 experiments, in Figure \ref{fig:zeta}-a and Figure \ref{fig:zeta}-b, we see that for $\zeta=0.4$, accuracy and NMI are indeed decreasing during the C3 training. This is because this value of $\zeta$ is too lenient and considers the points not in the same cluster to be similar. We can confirm this explanation by looking at Figure \ref{fig:zeta}-d. We can see that for smaller $\zeta$s, we will have more average positive pairs per sample. As we increase the $\zeta$, we can see that the performance begins to improve. For larger values such as $\zeta=0.9$, we can see that the performance does not significantly change during the training. This is because $\zeta = 0.9$ is a strict threshold, and if we look at the number of positive pairs, only a few instances are identified as similar during the training.
 
Comparing the results of CIFAR-10 and Tiny-ImageNet experiments shows that the value of $\zeta$ also depends on the number of clusters. Since we have 200 classes in Tiny-ImageNet, a smaller value of $\zeta$ might yield two or more clusters merging together and this would decrease the accuracy. Therefore, we should choose a more strict threshold such as $\zeta=0.9$ or $\zeta=0.95$ to improve. In Figure \ref{fig:zeta}-c and Figure \ref{fig:zeta}-g, the average loss plot also conveys interesting observations about the behaviour of $\zeta$. We can see that for smaller values, the loss is exponentially converging to the minimum, but for larger $\zeta$, the rate is much slower. This can be due to the fact that a smaller $\zeta$ considers most points to be similar and of the same class, and therefore, it can yield the trivial solution of considering all points to be similar and mapping them into one central point. In the extreme case of $\zeta \to 1$, C3 considers a few samples as positive pairs, and therefore, we will not have any major improvement. In contrast, if we set $\zeta \to -1$, the loss considers all points to be positive and the numerator and denominator of Eq. (\ref{C3loss}) become equal. Therefore, the loss function becomes zero and the network does not train. Following the above discussion, we suggest a value like $\zeta=0.6$, which is a good balance. However, the choice of $\zeta$ might be influenced by the number of clusters in the data. If we have a large number of clusters, it would be better to choose a large $\zeta$. On the other hand, if the data has a small number of clusters, a smaller $\zeta$ (but not too small) is preferred since it trains faster. In our experiments, we set $\zeta=0.6$ unless in Tiny-ImageNet which has 200 classes where we used $\zeta=0.95$.
\begin{wrapfigure}[8]{R}{0.5\textwidth}
    \centering
    \vspace{-8mm}
    \includegraphics[scale = 0.15]{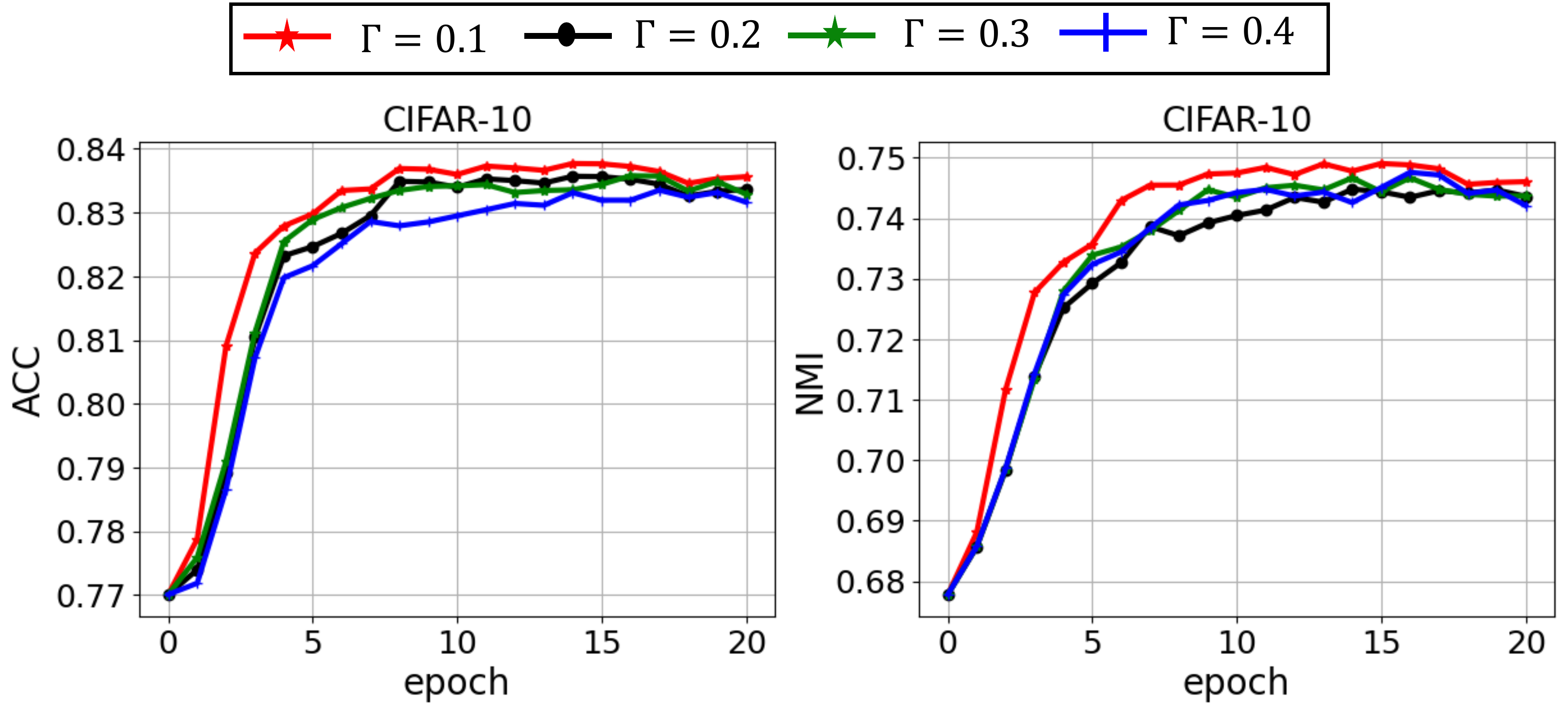}
  \caption{Performance of C3 for different values of $\Gamma$, for CIFAR 10.}
  \vspace{-5mm}
  \label{fig:gamma}
\end{wrapfigure}

Figure \ref{fig:gamma} illustrates the impact of hyperparameter $\Gamma$ on the performance C3. For very low values of $\Gamma$ (i.e., $\Gamma \xrightarrow[]{}0$), all weights converge to the same value of $w_{ij}^k=\frac{1}{2N}$. Conversely, for very high values of $\Gamma$ (i.e., $\Gamma \xrightarrow[]{} \infty$), the effect of entropy in Eq. \eqref{eq11} is neglected, leading to a trivial solution where our weighting function in Eq. \eqref{eq:weighting} selects one negative sample having the highest value of $1-|z_i^{a^\intercal} z_j^k|$ to minimize the first term in Eq. \eqref{eq11}. \s{WHAT??? Not clear. REVISE the following... one weight corresponding to the highest value of $1-|z_i^{a^\intercal} z_j^k|$ converges to one, and all other weights converge to zero. }\s{Our experiments show that the best value of $\Gamma$ for the CIFAR-10 dataset is 0.1}In all our experiments for all the datasets, $\Gamma$ is set to 0.1 though a better performance may be obtained if we fine-tune it per dataset.\s{ and we do not change its value for other datasets.} Overall, our results demonstrate the importance of selecting an appropriate value for $\Gamma$ to optimize the performance of our proposed method.
\vspace{-6mm}
\section{Conclusion}
\vspace{-3mm}
In this paper, we proposed C3, an algorithm for contrastive data clustering that incorporates the similarity between different instances to form a better representation for clustering. We experimentally showed that our method could significantly outperform the state-of-the-art on five challenging computer vision datasets. In addition, through additional experiments, we evaluated different aspects of our algorithm and provided several intuitions on how and why our proposed scheme can help in learning a more cluster-friendly representation. The focus of this work was on image clustering, but our idea can also be applied to clustering other data types, such as text and categorical data, in future works.

\bibliography{egbib}

\begin{thebibliography}{43}
\providecommand{\natexlab}[1]{#1}
\providecommand{\url}[1]{\texttt{#1}}
\expandafter\ifx\csname urlstyle\endcsname\relax
  \providecommand{\doi}[1]{doi: #1}\else
  \providecommand{\doi}{doi: \begingroup \urlstyle{rm}\Url}\fi

\bibitem[Bengio et~al.(2006)Bengio, Lamblin, Popovici, and Larochelle]{ae}
Yoshua Bengio, Pascal Lamblin, Dan Popovici, and Hugo Larochelle.
\newblock Greedy layer-wise training of deep networks.
\newblock In B.~Sch\"{o}lkopf, J.~Platt, and T.~Hoffman, editors,
  \emph{Advances in Neural Information Processing Systems}, volume~19. MIT
  Press, 2006.
\newblock URL
  \url{https://proceedings.neurips.cc/paper/2006/file/5da713a690c067105aeb2fae32403405-Paper.pdf}.

\bibitem[Cai et~al.(2009)Cai, He, Wang, Bao, and Han]{nmf}
Deng Cai, Xiaofei He, Xuanhui Wang, Hujun Bao, and Jiawei Han.
\newblock Locality preserving nonnegative matrix factorization.
\newblock In \emph{Proceedings of the 21st International Joint Conference on
  Artificial Intelligence}, IJCAI'09, page 1010–1015, San Francisco, CA, USA,
  2009. Morgan Kaufmann Publishers Inc.

\bibitem[Cai et~al.(2022)Cai, Fan, Guo, Wang, Zhang, and Zhang]{edesc}
Jinyu Cai, Jicong Fan, Wenzhong Guo, Shiping Wang, Yunhe Zhang, and Zhao Zhang.
\newblock Efficient deep embedded subspace clustering.
\newblock In \emph{2022 IEEE/CVF Conference on Computer Vision and Pattern
  Recognition (CVPR)}, pages 21--30, 2022.
\newblock \doi{10.1109/CVPR52688.2022.00012}.

\bibitem[Chang et~al.(2017)Chang, Wang, Meng, Xiang, and Pan]{dac}
Jianlong Chang, Lingfeng Wang, Gaofeng Meng, Shiming Xiang, and Chunhong Pan.
\newblock Deep adaptive image clustering.
\newblock In \emph{2017 IEEE International Conference on Computer Vision
  (ICCV)}, pages 5880--5888, 2017.
\newblock \doi{10.1109/ICCV.2017.626}.

\bibitem[Chang et~al.(2019)Chang, Guo, Wang, Meng, Xiang, and Pan]{DDC}
Jianlong Chang, Yiwen Guo, Lingfeng Wang, Gaofeng Meng, Shiming Xiang, and
  Chunhong Pan.
\newblock Deep discriminative clustering analysis.
\newblock \emph{ArXiv}, abs/1905.01681, 2019.

\bibitem[Chen et~al.(2020)Chen, Kornblith, Norouzi, and Hinton]{SimCLR}
Ting Chen, Simon Kornblith, Mohammad Norouzi, and Geoffrey Hinton.
\newblock A simple framework for contrastive learning of visual
  representations.
\newblock \emph{arXiv preprint arXiv:2002.05709}, 2020.

\bibitem[Deng et~al.()Deng, Dong, Socher, Li, Li, and Fei-Fei]{Imagenet}
Jia Deng, Wei Dong, Richard Socher, Li-Jia Li, Kai Li, and Li~Fei-Fei.
\newblock Imagenet full (fall 2011 release).

\bibitem[Feng et~al.(2019)Feng, Xu, and Tao]{Rotate}
Zeyu Feng, Chang Xu, and Dacheng Tao.
\newblock Self-supervised representation learning by rotation feature
  decoupling.
\newblock In \emph{2019 IEEE/CVF Conference on Computer Vision and Pattern
  Recognition (CVPR)}, pages 10356--10366, 2019.
\newblock \doi{10.1109/CVPR.2019.01061}.

\bibitem[Gowda and Krishna(1978)]{AC}
K.~Chidananda Gowda and G.~Krishna.
\newblock Agglomerative clustering using the concept of mutual nearest
  neighbourhood.
\newblock \emph{Pattern Recognit.}, 10:\penalty0 105--112, 1978.

\bibitem[Guo et~al.(2017)Guo, Gao, Liu, and Yin]{IDEC}
Xifeng Guo, Long Gao, Xinwang Liu, and Jianping Yin.
\newblock Improved deep embedded clustering with local structure preservation.
\newblock In \emph{Proceedings of the 26th International Joint Conference on
  Artificial Intelligence}, IJCAI'17, page 1753–1759. AAAI Press, 2017.
\newblock ISBN 9780999241103.

\bibitem[Guo et~al.(2018)Guo, Zhu, Liu, and Yin]{DECDA}
Xifeng Guo, En~Zhu, Xinwang Liu, and Jianping Yin.
\newblock Deep embedded clustering with data augmentation.
\newblock In Jun Zhu and Ichiro Takeuchi, editors, \emph{Proceedings of The
  10th Asian Conference on Machine Learning}, volume~95 of \emph{Proceedings of
  Machine Learning Research}, pages 550--565. PMLR, 14--16 Nov 2018.
\newblock URL \url{https://proceedings.mlr.press/v95/guo18b.html}.

\bibitem[Haeusser et~al.(2019)Haeusser, Plapp, Golkov, Aljalbout, and
  Cremers]{adc}
Philip Haeusser, Johannes Plapp, Vladimir Golkov, Elie Aljalbout, and Daniel
  Cremers.
\newblock Associative deep clustering: Training a classification network with
  no labels.
\newblock In Thomas Brox, Andr{\'e}s Bruhn, and Mario Fritz, editors,
  \emph{Pattern Recognition}, pages 18--32, Cham, 2019. Springer International
  Publishing.
\newblock ISBN 978-3-030-12939-2.

\bibitem[He et~al.(2016)He, Zhang, Ren, and Sun]{resnet}
Kaiming He, Xiangyu Zhang, Shaoqing Ren, and Jian Sun.
\newblock Deep residual learning for image recognition.
\newblock In \emph{Proceedings of the IEEE conference on computer vision and
  pattern recognition}, pages 770--778, 2016.

\bibitem[Hojjati et~al.(2022)Hojjati, Ho, and
  Armanfard]{Hojjati2022SelfSupervisedAD}
Hadi Hojjati, Thi Kieu~Khanh Ho, and Narges Armanfard.
\newblock Self-supervised anomaly detection: A survey and outlook.
\newblock \emph{ArXiv}, abs/2205.05173, 2022.

\bibitem[Huang et~al.(2020)Huang, Gong, and Zhu]{PICA}
Jiabo Huang, Shaogang Gong, and Xiatian Zhu.
\newblock Deep semantic clustering by partition confidence maximisation.
\newblock In \emph{Proceedings of the IEEE/CVF Conference on Computer Vision
  and Pattern Recognition (CVPR)}, June 2020.

\bibitem[Huang et~al.(2014)Huang, Huang, Wang, and Wang]{DEN}
Peihao Huang, Yan Huang, Wei Wang, and Liang Wang.
\newblock Deep embedding network for clustering.
\newblock In \emph{2014 22nd International Conference on Pattern Recognition},
  pages 1532--1537, 2014.
\newblock \doi{10.1109/ICPR.2014.272}.

\bibitem[Ji et~al.(2019)Ji, Henriques, and Vedaldi]{IIC}
Xu~Ji, Jo{\~a}o~F Henriques, and Andrea Vedaldi.
\newblock Invariant information clustering for unsupervised image
  classification and segmentation.
\newblock In \emph{Proceedings of the IEEE International Conference on Computer
  Vision}, pages 9865--9874, 2019.

\bibitem[Kingma and Welling(2014)]{vae}
Diederik~P. Kingma and Max Welling.
\newblock Auto-encoding variational bayes.
\newblock \emph{CoRR}, abs/1312.6114, 2014.

\bibitem[Krizhevsky et~al.()Krizhevsky, Nair, and Hinton]{Cifar}
Alex Krizhevsky, Vinod Nair, and Geoffrey Hinton.
\newblock Cifar-10 (canadian institute for advanced research).
\newblock URL \url{http://www.cs.toronto.edu/~kriz/cifar.html}.

\bibitem[Kullback and Leibler(1951)]{KL}
S.~Kullback and R.~A. Leibler.
\newblock On information and sufficiency.
\newblock \emph{Ann. Math. Statist.}, 22\penalty0 (1):\penalty0 79--86, 1951.

\bibitem[Le and Yang(2015)]{Tinyimage}
Ya~Le and Xuan~S. Yang.
\newblock Tiny imagenet visual recognition challenge, 2015.

\bibitem[Li et~al.(2021)Li, Hu, Liu, Peng, Zhou, and Peng]{CC}
Yunfan Li, Peng Hu, Zitao Liu, Dezhong Peng, Joey~Tianyi Zhou, and Xi~Peng.
\newblock Contrastive clustering.
\newblock \emph{Proceedings of the AAAI Conference on Artificial Intelligence},
  35\penalty0 (10):\penalty0 8547--8555, May 2021.
\newblock \doi{10.1609/aaai.v35i10.17037}.
\newblock URL \url{https://ojs.aaai.org/index.php/AAAI/article/view/17037}.

\bibitem[Lin et~al.(2021)Lin, Qi, Zhengyang, and Changhu]{lin2021inter}
Z~Lin, S~Qi, S~Zhengyang, and W~Changhu.
\newblock Inter-intra variant dual representations for self-supervised video
  recognition.
\newblock In \emph{British Machine Vision Conference}, volume~2, page~7, 2021.

\bibitem[Lloyd(1982)]{kmeans}
Stuart~P. Lloyd.
\newblock Least squares quantization in pcm.
\newblock \emph{IEEE Trans. Inf. Theory}, 28:\penalty0 129--136, 1982.

\bibitem[Niu et~al.(2020)Niu, Zhang, Wang, and Liang]{gatcluster2020}
Chuang Niu, Jun Zhang, Ge~Wang, and Jimin Liang.
\newblock Gatcluster: Self-supervised gaussian-attention network for image
  clustering.
\newblock In \emph{European Conference on Computer Vision (ECCV)}, 2020.

\bibitem[Peng et~al.(2016)Peng, Xiao, Feng, Yau, and Yi]{PARTY}
Xi~Peng, Shijie Xiao, Jiashi Feng, Wei-Yun Yau, and Zhang Yi.
\newblock Deep subspace clustering with sparsity prior.
\newblock In \emph{Proceedings of the Twenty-Fifth International Joint
  Conference on Artificial Intelligence}, IJCAI'16, page 1925–1931. AAAI
  Press, 2016.
\newblock ISBN 9781577357704.

\bibitem[Radford et~al.(2016)Radford, Metz, and Chintala]{dcgan}
Alec Radford, Luke Metz, and Soumith Chintala.
\newblock Unsupervised representation learning with deep convolutional
  generative adversarial networks.
\newblock \emph{CoRR}, abs/1511.06434, 2016.

\bibitem[Sadeghi and Armanfard(2021{\natexlab{a}})]{DCSS}
Mohammadreza Sadeghi and Narges Armanfard.
\newblock {Deep Clustering with Self-supervision using Pairwise Data
  Similarities}.
\newblock \emph{TechRxiv}, 6 2021{\natexlab{a}}.
\newblock \doi{10.36227/techrxiv.14852652.v1}.
\newblock URL
  \url{https://www.techrxiv.org/articles/preprint/Deep_Clustering_with_Self-supervision_using_Pairwise_Data_Similarities/14852652}.

\bibitem[Sadeghi and Armanfard(2021{\natexlab{b}})]{IDECF}
Mohammadreza Sadeghi and Narges Armanfard.
\newblock Idecf: Improved deep embedding clustering with deep fuzzy
  supervision.
\newblock In \emph{2021 IEEE International Conference on Image Processing
  (ICIP)}, pages 1009--1013, 2021{\natexlab{b}}.
\newblock \doi{10.1109/ICIP42928.2021.9506051}.

\bibitem[Sadeghi and Armanfard(2022)]{DML}
Mohammadreza Sadeghi and Narges Armanfard.
\newblock {Deep Multi-Representation Learning for Data Clustering}.
\newblock \emph{TechRxiv}, 3 2022.
\newblock \doi{10.36227/techrxiv.19357979.v1}.
\newblock URL
  \url{https://www.techrxiv.org/articles/preprint/Deep_Multi-Representation_Learning_for_Data_Clustering/19357979}.

\bibitem[Shiran and Weinshall(2020)]{MMDC}
Guy Shiran and Daphna Weinshall.
\newblock Multi-modal deep clustering: Unsupervised partitioning of images.
\newblock In \emph{International Conference on Pattern Recognition (ICPR)},
  2020.

\bibitem[Tiwari et~al.(2021)Tiwari, Chen, Tsai, Kuo, Chen, Jou, Venkatesh, and
  Chen]{tiwari2021self}
Hitika Tiwari, Min-Hung Chen, Yi-Min Tsai, Hsien-Kai Kuo, Hung-Jen Chen, Kevin
  Jou, KS~Venkatesh, and Yong-Sheng Chen.
\newblock Self-supervised robustifying guidance for monocular 3d face
  reconstruction.
\newblock 2021.

\bibitem[Van~Gansbeke et~al.(2020)Van~Gansbeke, Vandenhende, Georgoulis,
  Proesmans, and Van~Gool]{SCAN}
Wouter Van~Gansbeke, Simon Vandenhende, Stamatios Georgoulis, Marc Proesmans,
  and Luc Van~Gool.
\newblock Scan: Learning to classify images without labels.
\newblock In \emph{Proceedings of the European Conference on Computer Vision},
  2020.

\bibitem[Vincent et~al.(2010)Vincent, Larochelle, Lajoie, Bengio, and
  Manzagol]{dae}
Pascal Vincent, Hugo Larochelle, Isabelle Lajoie, Yoshua Bengio, and
  Pierre-Antoine Manzagol.
\newblock Stacked denoising autoencoders: Learning useful representations in a
  deep network with a local denoising criterion.
\newblock \emph{Journal of Machine Learning Research}, 11\penalty0
  (110):\penalty0 3371--3408, 2010.
\newblock URL \url{http://jmlr.org/papers/v11/vincent10a.html}.

\bibitem[Wu et~al.(2019)Wu, Long, Wang, Qian, Li, Lin, and Zha]{DCCM}
Jianlong Wu, Keyu Long, Fei Wang, Chen Qian, Cheng Li, Zhouchen Lin, and
  Hongbin Zha.
\newblock Deep comprehensive correlation mining for image clustering.
\newblock In \emph{International Conference on Computer Vision}, 2019.

\bibitem[Xie et~al.(2021)Xie, Ding, Wang, Zhan, Xu, Sun, Li, and
  Luo]{Objectrecognition}
Enze Xie, Jian Ding, Wenhai Wang, Xiaohang Zhan, Hang Xu, Peize Sun, Zhenguo
  Li, and Ping Luo.
\newblock Detco: Unsupervised contrastive learning for object detection.
\newblock In \emph{2021 IEEE/CVF International Conference on Computer Vision
  (ICCV)}, pages 8372--8381, 2021.
\newblock \doi{10.1109/ICCV48922.2021.00828}.

\bibitem[Xie et~al.(2016)Xie, Girshick, and Farhadi]{DEC}
Junyuan Xie, Ross Girshick, and Ali Farhadi.
\newblock Unsupervised deep embedding for clustering analysis.
\newblock In \emph{Proceedings of the 33rd International Conference on
  International Conference on Machine Learning - Volume 48}, ICML'16, page
  478–487. JMLR.org, 2016.

\bibitem[Yang et~al.(2017)Yang, Fu, Sidiropoulos, and Hong]{DCN}
Bo~Yang, Xiao Fu, Nicholas~D. Sidiropoulos, and Mingyi Hong.
\newblock Towards k-means-friendly spaces: Simultaneous deep learning and
  clustering.
\newblock In \emph{Proceedings of the 34th International Conference on Machine
  Learning - Volume 70}, ICML'17, page 3861–3870. JMLR.org, 2017.

\bibitem[Yang et~al.(2016)Yang, Parikh, and Batra]{JULE}
J.~Yang, D.~Parikh, and D.~Batra.
\newblock Joint unsupervised learning of deep representations and image
  clusters.
\newblock In \emph{2016 IEEE Conference on Computer Vision and Pattern
  Recognition (CVPR)}, pages 5147--5156, Los Alamitos, CA, USA, jun 2016. IEEE
  Computer Society.
\newblock \doi{10.1109/CVPR.2016.556}.
\newblock URL \url{https://doi.ieeecomputersociety.org/10.1109/CVPR.2016.556}.

\bibitem[Zeiler et~al.(2010)Zeiler, Krishnan, Taylor, and Fergus]{decnn}
Matthew~D. Zeiler, Dilip Krishnan, Graham~W. Taylor, and Rob Fergus.
\newblock Deconvolutional networks.
\newblock In \emph{2010 IEEE Computer Society Conference on Computer Vision and
  Pattern Recognition}, pages 2528--2535, 2010.
\newblock \doi{10.1109/CVPR.2010.5539957}.

\bibitem[Zelnik-manor and Perona(2004)]{SC}
Lihi Zelnik-manor and Pietro Perona.
\newblock Self-tuning spectral clustering.
\newblock In L.~Saul, Y.~Weiss, and L.~Bottou, editors, \emph{Advances in
  Neural Information Processing Systems}, volume~17. MIT Press, 2004.
\newblock URL
  \url{https://proceedings.neurips.cc/paper/2004/file/40173ea48d9567f1f393b20c855bb40b-Paper.pdf}.

\bibitem[Zhong et~al.(2020)Zhong, Chen, Jin, and Hua]{DRC}
Huasong Zhong, C.~Chen, Zhongming Jin, and Xiansheng Hua.
\newblock Deep robust clustering by contrastive learning.
\newblock \emph{ArXiv}, abs/2008.03030, 2020.

\bibitem[Zhou et~al.(2022)Zhou, Xu, Zheng, Chen, Li, Bu, Wu, Wang, Zhu, and
  Ester]{reviewCluster}
Sheng Zhou, Hongjia Xu, Zhuonan Zheng, Jiawei Chen, Zhao Li, Jiajun Bu, Jia Wu,
  Xin Wang, Wenwu Zhu, and Martin Ester.
\newblock A comprehensive survey on deep clustering: Taxonomy, challenges, and
  future directions.
\newblock \emph{ArXiv}, abs/2206.07579, 2022.

\end{thebibliography}
\end{document}